\documentclass[journal]{IEEEtran}
\usepackage{amsmath,amsfonts}
\usepackage{algorithmic}
\usepackage{algorithm}
\usepackage{array}
\usepackage[caption=false,font=normalsize,labelfont=sf,textfont=sf]{subfig}
\usepackage{textcomp}
\usepackage{stfloats}
\usepackage{url}
\usepackage{verbatim}
\usepackage{graphicx}
\usepackage{cite}
\usepackage{booktabs}
\usepackage{multirow}
\usepackage{xcolor}
\usepackage{color}
\usepackage{makecell}
\begin{document}

\title{An Inertial Sequence Learning Framework for Vehicle Speed Estimation via Smartphone IMU}

\author{Xuan~Xiao, Xiaotong~Ren, Haitao~Li

    \thanks{}
}


\maketitle

\begin{abstract}
    Accurately estimating vehicle velocity via smartphone is critical for mobile navigation and transportation. This paper introduces a cutting-edge framework for velocity estimation that incorporates temporal learning models, utilizing Inertial Measurement Unit (IMU) data and is supervised by Global Navigation Satellite System (GNSS) information. The framework employs a noise compensation network to fit the noise distribution between sensor measurements and actual motion, and a pose estimation network to align the coordinate systems of the phone and the vehicle. To enhance the model's generalizability, a data augmentation technique that mimics various phone placements within the car is proposed. Moreover, a new loss function is designed to mitigate timestamp mismatches between GNSS and IMU signals, effectively aligning the signals and improving the velocity estimation accuracy. Finally, we implement a highly efficient prototype and conduct extensive experiments on a real-world crowdsourcing dataset, resulting in superior accuracy and efficiency.
\end{abstract}


\section{Introduction}
\begin{figure}[tbp]
    \centerline{\includegraphics[width=0.5\textwidth]{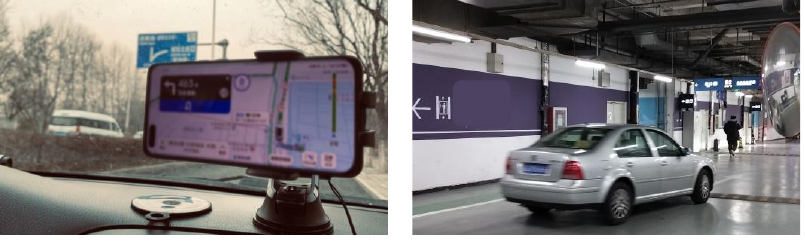}}
    \caption{Left: Driving with the assistance of navigation application. Right: Vehicle drving in GNSS-blocked environments.}
    \label{fig:intro}
\end{figure}

\IEEEPARstart{T}{he} 
emergence of smartphone-based vehicular applications has revolutionized how drivers access and take advantage of mobile services. These applications offer a wide range of valuable features that enhance driving safety and convenience, such as real-time vehicle positioning, analysis of driving behavior, intelligent navigation assistance, and traffic status updates. 
According to statistics, in 2021, nearly 70\% of drivers use mobile navigation apps like Gaode and Baidu Maps while driving (Fig. \ref{fig:intro} (a)) in China. Ride-hailing drivers, in particular, rely heavily on the positioning services provided by these mobile navigation apps to ensure accurate passenger pick-up and drop-off. Consequently, navigation app service providers, such as DiDi, Uber, and Amap, are dedicated to enhancing the precision of smartphone-based vehicle positioning, thereby improving the user experience.

Typically, Global Navigation Satellite System (GNSS) information provides position \cite{boguspayev2023comprehensive}.
However, the limitations of mobile phone hardware and complex urban environments can lead to signal degradation and even congestion, which challenges GNSS to provide a consistently stable signal over long periods of time, especially when the vehicle passes through densely built areas, tunnels, or underground parking facilities (Fig. \ref{fig:intro} (b)). 
The absence of satellite perception significantly hampers the driving experience, for instance, in subterranean parking lots where the provided location diverges considerably from the actual position, driver may encounter confusion and disorientation.

However, apart from GNSS, there are other ways to offer mobile-based vehicle positioning services.
A couple of alternatives based on On-Board Diagnostics (OBD) or Radio Frequency (RF) signal. OBD-II can retrieve vehicle speeds, but requires an adapter, and RF signal such as Wi-Fi based methods\cite{wifi2023track} require additional device deployment, making them not widely available.
For navigation application service providers, the need for extra equipment hampers application promotion. They prefer to offer accurate vehicle tracking using only the mobile phone's built-in sensors, without any noticeable impact on the user. Consequently, the built-in Inertial Measurement Unit (IMU) in smartphone has gained immense popularity among academics and industry \cite{brossard2020ai,gao2021glow,zhou2022deepvip} as they don't require any external infrastructure and consume low power. 


\IEEEpubidadjcol

In our prior work \cite{tong2022vehicle}, we identified three key challenges in smartphone IMU-based vehicle positioning through the observation and analysis of numerous navigation trajectories: 1) Inconstant Inertial Errors: the inertial errors vary dramatically through crowdsensing; 2) Arbitrary Posture of Smartphones: drivers have their own preferences for positioning their smartphones; and 3) The Diversity of Hardware: the diverse range of commodity inertial sensors from different manufacturers results in varying localization accuracies. These challenges render traditional filtering methods, such as EKF \cite{ekftits}, ineffective, particularly due to the uncertainty in smartphone posture, which frequently prevents EKF from converging effectively. Therefore, we adopted deep learning-based approaches \cite{gao2021glow,tong2022vehicle}. Deep learning leverages the benefits of big data to model the dynamic motion of the IMU, thereby facilitating more stable vehicle position tracking. Moreover, deep learning-based methods have a higher ceiling and offer potential for improvement. There is considerable room for enhancement in the previous works. Firstly, the model's generalization performance requires enhancement, as general models continue to degrade when applied across different mobile phones. Secondly, the model's capacity to fit data needs improvement. High-quality training data is inadequate to cover a wide range of application scenarios, while low-quality data challenges the model's accuracy; Finally, there is a need to enhance efficiency, with mobile navigation favoring rapid and energy-efficient tracking algorithms.

In this paper, we present a novel vehicle speed estimation method based on sequence learning, called DVSE (Deep learning-based Vehicle Speed Estimation). DVSE utilizes the IMU data from smartphones to estimate the vehicle's speed. Unlike traditional methods that require complex initialization, DVSE is designed to be plug-and-play, requiring no additional setup time. Our method enables flexible smartphone placement inside the vehicle, providing convenience and practicality in real-world scenarios. Moreover, We train our method using GNSS data as the supervised source, eliminating the need for expensive and cumbersome ground truth data collection devices.

We highlight our main contributions as follows:
\begin{itemize}
    \item We propose a novel framework that utilizes temporal learning models to estimate vehicle velocity. By fitting the noise model to the overall noise distribution between sensor measurements and actual motion, we simplify the learning objective of the framework, thereby enhancing its robustness and improving speed estimation accuracy. 
    \item We explore a motion transformation network that aligns the coordinate systems of the smartphone and the vehicle during the velocity estimation process. To improve the generalizability of the model, we develop a data augmentation method that simulates various placements of the smartphone inside the vehicle, thereby minimizing the influence of varied phone positioning postures on the system.
    \item We design a unique matching loss function that tackles the problem of mismatched timestamps between GNSS and IMU signals, resulting in enhanced model accuracy and robustness while simplifying the filtering of training data.
    \item We develop a highly efficient prototype and conduct extensive experiments on a real-world crowdsourcing dataset. Results have shown the effectiveness of our proposed method.
\end{itemize}

The following of this paper is organized as follows:

We firstly introduct the related work in Section \ref{sec:related} and then demonstrate our system architecture in Section \ref{sec:overview}. 
In section \ref{sec:noise}, we describe the noise compensation network.
The pose transformation network is presented in Section \ref{sec:mtn}.
Section \ref{sec:loss} shows how we calculate the loss of the model, and Section \ref{sec:exp} illustrates our experimental design, results, and analysis. Section \ref{sec:conclusion} conclude the paper.

\section{Related Work} \label{sec:related}
In this section, we review the existing works on vehicle speed estimation, motion transformation and sequential deep Learning.
\subsection{Vehicle Speed Estimation}
We categories the vehicle speed estimation into external devices-based and IMU-based. External devices specifically refer to sensor devices that are not internally present in smartphones, while IMU refers to the low-cost IMU embedded in smartphones.
\subsubsection{External Devices-based}
Previous studies have explored vehicle speed estimation by utilizing external devices such as vehicle control unit (VCU) \cite{ding2020vse}, OBD-II adapter \cite{wang2013sensing}, radar \cite{radar}, traffic magnetic sensors \cite{li2011some,magtits} or UWB \cite{uwbtits}. These approaches enable high-precision estimation of vehicle speed but relies entirely on independent devices. 
Other popular methods, such as wifi \cite{wifi2018track, wifi2023track}, allow for accurate vehicle positioning without cumulative errors. However, despite the signals being accessible via smartphones, they still depend on external signal-generating devices.
While external devices-based methods can provide accurate results, the reliance on additional deployments limits their practicality and scalability.
\subsubsection{IMU-based}
In recent years, there has been a growing interest in leveraging IMU for vehicle speed estimation, especially for scenarios where GNSS fails.
Traditional methods use filters to model the motion dynamic, such as Kalman and particle filters\cite{doumiati2012vehicle, yu2022tightly, gao2017smartphone}. While effective, they often suffer from drift and require careful calibration and tuning to achieve accurate results. 
With the recent advancements in deep learning, researchers have explored using neural networks for inertial tracking \cite{chen2018ionet, herath2020ronin, brossard2020ai, li2022learning}. These methods leverage the power of deep learning models to learn complex patterns and relationships in sensor data.
In addition, some data-driven works, such as \cite{gao2021glow, tong2022vehicle, zhou2022deepvip}, have shown promising results in improving the accuracy and robustness by training on large datasets.
These methods offer the advantage of using existing smartphone hardware, making them more accessible and cost-effective.  

\subsection{Motion Transformation}
Using the motion readings of the IMU in smartphone to observe the dynamics of the vehicle, it is inevitable to align the two coordinate systems.
One approach is to transform both systems into the world coordinate system, often utilizing geomagnetic information \cite{zhou2022deepvip}. However, relying on the smartphone's embedded magnetometer is susceptible to interference from the magnetic field in the surrounding environment, especially in the presence of magnetic disturbances on the vehicle.
Another approach is to solve the relative pose between a smartphone and a vehicle through IMU measurements, which relies on two key clues: the alignment of the gravity acceleration, and the correlation between the linear acceleration (excluding gravity) of the smartphone and the heading direction of the vehicle when moving along a straight line.
The gravity acceleration measurement is straightforward, but determining the forward direction is more challenging. 
Some methods align the smartphone coordinate system with the vehicle coordinate system using techniques such as projection \cite{gao2017smartphone} or principal component analysis (PCA) \cite{tong2022vehicle}. These methods rely on an initialization to determine the forward direction, which means any changes in the smartphone's pose due to vibrations or touch during the driving can lead to an incorrecttion of the initialized rotation angles.

\subsection{Sequential Deep Learning}
Several models have proven effective in accomplishing temporal relationship modeling, such as RNN \cite{yu2019review} (LSTM \cite{hochreiter1997long}, GRU \cite{chung2014empirical}), TCN \cite{bai2018empirical}, and Attention \cite{vaswani2017attention}. Attention has garnered significant attention lately due to its exceptional performance in large language models. It exhibits strong modeling capabilities for sequential data \cite{zhou2021informer}. However, its structure is relatively complex and requires a considerable number of parameters to maintain good performance. LSTM and GRU belong to the family of recurrent neural networks and are commonly employed in various time series modeling problems. GRU has been empirically demonstrated to achieve comparable precision to LSTM while utilizing a smaller parameter count in specific scenarios. TCN, on the other hand, is a lightweight model that effectively captures temporal patterns in data.

\section{Overview} \label{sec:overview}
In this section, we present the design rationale behind the DVSE framework and provide a comprehensive description of the system architecture.
\subsection{Motion Model}
We first analyze the relationship between mobile phone sensors and vehicle speed. The IMU consists of an accelerometer and a gyroscope. The accelerometer measures the acceleration produced by the forces acting on the smartphone and the sensor provides observations along three axes, corresponding to the three directions of the smartphone. Specifically, the accelerometer measurements from smartphone's IMU, $\hat{\boldsymbol{\alpha} _p}$, is given by:
\begin{equation}
    \hat{\boldsymbol{\alpha} _p}=\boldsymbol{\alpha}_p+\boldsymbol{R}^{p}_w \boldsymbol{g}^w +\boldsymbol{N}_{\alpha}
\end{equation}
where $\boldsymbol{g}^w$ represents the gravitational acceleration, $\boldsymbol{R}^{p}_w$ denotes the rotation from the world frame to the smartphone frame, and $\boldsymbol{\alpha}_p$ represents the linear acceleration excluding the gravitational component. Furthermore, the observations are also affected by the bias inherent to the IMU and the measurement noise, which are donated to $\boldsymbol{N}$.

\begin{figure}[htbp]
    \centerline{\includegraphics[width=0.45\textwidth]{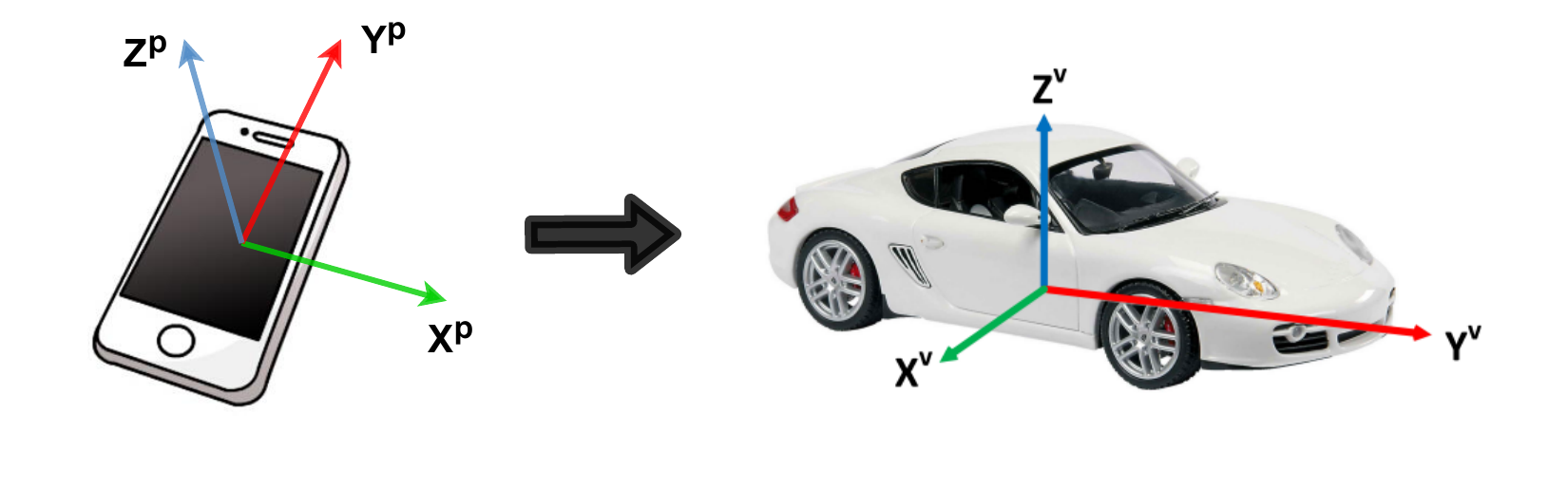}}
    \caption{Coordinate system transformation from the phone to the vehicle.}
    \label{fig:axes_trans}
\end{figure}

When the smartphone is relatively fixed to the vehicle, the smartphone's motion can be utilized to capture the vehicle's movement, as illustrated in Fig. \ref{fig:axes_trans}. The linear acceleration of the car can be expressed as:
\begin{equation}
    \boldsymbol{\alpha}_v=\boldsymbol{R}^v_p \boldsymbol{\alpha}_p=\boldsymbol{R}^v_p(\hat{\boldsymbol{\alpha }}_p-\boldsymbol{R}^p_w \boldsymbol{g}^w-\boldsymbol{N}_{\alpha})
\end{equation}
Therefore, the propagation model of the vehicle's speed within the time interval $\Delta t$ can be defined as:
\begin{equation}\label{eq:v}
    \boldsymbol{v}_{t+\Delta t} = \boldsymbol{v}_t+ \Delta \boldsymbol{v}=\boldsymbol{v}_t+
    \int_{t\in [t,t+\Delta t]} \boldsymbol{\alpha}_v \,dt
\end{equation}
Obviously, noise is one of the critical factor that affects accuracy. 
However, for consumer-grade smartphone IMUs, it is challenging to establish a unified noise model that can adapt to all scenarios. The noise characteristics may vary in contexts, even for the same device. In addition, (\ref{eq:v}) calculates the car's velocity along all three axes, indicating the need to consider noise and bias for each axis.

Fortunately, in practical applications, we are primarily concerned with the forward velocity of the vehicle, which is the velocity in the Y-axis direction. When we only need to calculate the forward velocity of the car, we can simplify the computational logic. The propagation equation for the forward velocity $v^{y}$ of the vehicle within $\Delta t$ can be defined as:
\begin{equation}
    v^{y}_{t+\Delta t} = v^{y}_t + \Delta v^y
\end{equation}
where,
\begin{equation} \label{eq:vy}
    \begin{aligned}
    \Delta v^y &= \int_{t\in [t,t+\Delta t]} \boldsymbol{\alpha}_v \cdot \boldsymbol{u} \,dt \\
    &=\int_{t\in [t,t+\Delta t]} [\boldsymbol{R}^v_p(\hat{\boldsymbol{\alpha}}_p-\boldsymbol{R}^p_w \boldsymbol{g}^w-\boldsymbol{N}_{\alpha})] \cdot \boldsymbol{u} \,dt\\
    &=\int_{t\in [t,t+\Delta t]} [\boldsymbol{R}^v_p(\hat{\boldsymbol{\alpha}}_p-\boldsymbol{R}^p_w \boldsymbol{g}^w)] \cdot \boldsymbol{u} \,dt + {\mathcal{N}}\\
    \end{aligned}
\end{equation}
\begin{equation}
    \boldsymbol{u} = [0,1,0]^T
\end{equation}
We distinguish between the integration term of acceleration and the disturbance term, represented by ${\mathcal{N}}$, which encompasses errors due to bias and noise. While the integration term is well-defined, accurately modeling the error term presents challenges. By treating these terms separately, the problem can be formulated as determining the optimal ${\mathcal{N}}$ that minimizes the disparity between the calculated result and the ground truth.

It is worth mentioning that by solely focusing on calculating the forward velocity of the vehicle, not only simplify the computation, but it also provides a premise for using GNSS as the ground truth. The speed information provided by GNSS is derived from the difference in displacement between two consecutive position points. 
Although GNSS measurements have inherent errors, their accuracy is generally sufficient for practical applications, and they are easily accessible. Therefore, we use GNSS as the target for learning the motion sequence of the IMU. 

\subsection{Model Architecture}
\begin{figure}[htbp]
    \centerline{\includegraphics[width=0.5\textwidth]{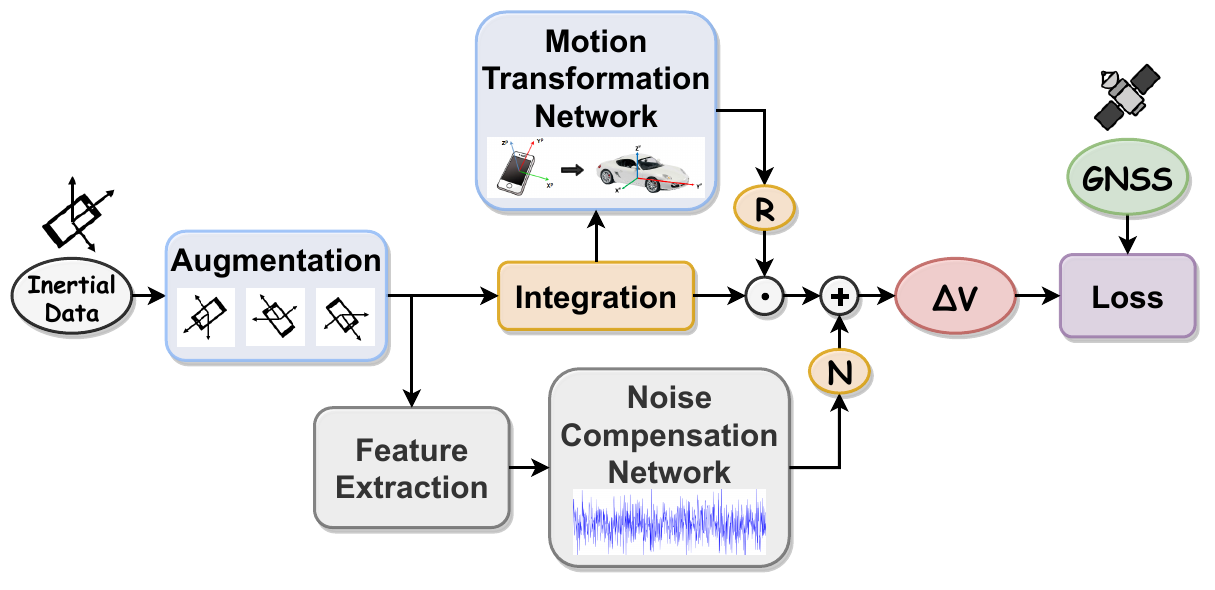}}
    \caption{The architecture of our model. The model consists three parts: sequential learning for noise (Sec. \ref{sec:noise}), motion transformation from phone to vehicle (Sec. \ref{sec:mtn}), and loss calculation (Sec. \ref{sec:loss}). $\bigodot $ stands for dot multiplication, and $\bigoplus $ stands for addition.}
    \label{fig:overview}
\end{figure}

For the sequence learning task, we take a sequence as the input and output corresponding sequence. The goal of inertial sequence learning is to find a network $f:x\Rightarrow y$ to minimize the difference between the estimation motion and the actual motion of the vehicle. With the aid of the modeling capability of deep learning, we can fit a function that transforms the phone's inertial data within $\Delta t$ into the corresponding vehicle's velocity deviation. Note that this estimated result contains no future information, i.e., we do not predict speed for next second but estimate speed current second.

The architecture of DVSE shows in Fig. \ref{fig:overview}. After data augmentation, the smartphone inertial data undergo two distinct operations: 1) The data is subjected to manual feature extraction, followed by the fitting of noise compensation. 2) The relative pose between the smartphone and the vehicle is calculated, leading to the transformation of the input data accordingly. Subsequently, the transformed motion measurements are combined with the estimated noise to generate the final vehicle velocity change. Finally, the GNSS data is employed as the ground truth to calculate the loss.

In our previous work \cite{gao2021glow,tong2022vehicle}, we have employed a holistic approach by allowing the model to learn the entire modeling process of (\ref{eq:vy}), which includes multiple logical steps such as coordinate system rotation, integration, and noise fitting within a time series model. However, this approach undoubtedly burdens the model's learning process. The effectiveness of the model deteriorates when there is insufficient training data or poor data quality. As evident from Equation (\ref{eq:vy}), if we solve for the relative pose between the smartphone and the vehicle ($\boldsymbol{R}^v_p$), as well as the observation errors (${\mathcal{N}}$), we can directly calculate the velocity. Therefore, we utilize different models to fit these two processes separately, which enables the models to learn specific tasks more effectively, thereby enhancing the model's learning performance and generalization capabilities.


\section{Noise Compensation} \label{sec:noise}
In this section, we utilize a sequence learning model to solve for the ${\mathcal{N}}$ terms in Equation \ref{eq:vy}. We begin by providing a brief analysis of the noise present in the sensors. Following that, we construct a learning model to fit this noise, and finally, we explain the input data for the model.
\subsection{Analysis of ${\mathcal{N}}$}
An IMU noise model often incorporates two categories: deterministic and stochastic \cite{nirmal2016noise}. 
Deterministic noises often arise from misalignment during the manufacturing process of sensor chips, static biases, or scale factor errors, which can be calibrated beforehand. On the other hand, stochastic errors such as quantization noise, random walk, flicker noise cannot be pre-calibrated. The typical approach is to approximate these noises with probability distributions such as Gaussian noise or white noise. However, in practical application scenarios, especially with low-cost devices, even pre-calibrated deterministic noises can deviate due to motion and temperature effects \cite{nikolic2015maximum}. 
Stochastic errors are even more challenging to determine. It is difficult to use a single generic noise distribution to approximate all possible types of noise. For smartphone sensors, especially in crowdsourcing scenarios, it is nearly impossible to pre-calibrate the sensors. 
Additionally, unforeseen disturbances such as irregular engine vibrations from the vehicle or user interference are also challenges to vehicle speed estimation processes.

Therefore, we utilize sequence learning to fit this demanding procedure.
With the data-driven and sequence learning model, we aim to learn the complex motion changes of the smartphone and account for sequential dependencies, which are very difficult to model by hand.
Note that our aim here is not to use a model to fit the noise distribution for a particular sensor, but merely to derive a general noise model suitable for smartphone-based vehicle speed estimation. 
\subsection{Model Design}
\begin{figure}[htbp]
    \centerline{\includegraphics[width=0.45\textwidth]{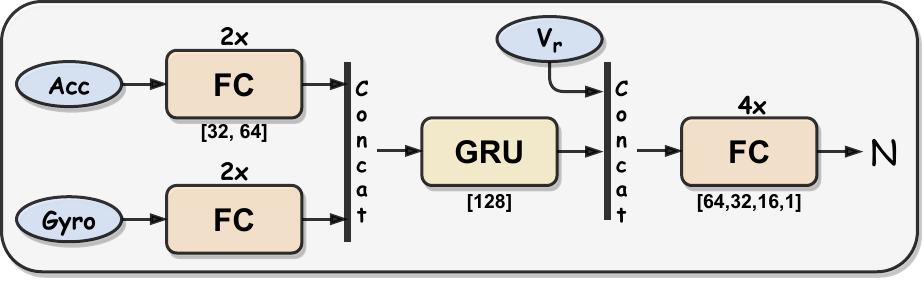}}
    \caption{Model architecture of the noise compensation block. 2x means that the module is repeated twice, and [32,64] means the number of hidden units.}
    \label{fig:noise_net}
\end{figure}

We employ the gated recurrent unit (GRU \cite{chung2014empirical}) as the core of model as its recurrent structure is well-suited for the noise characteristics. GRU uses gating mechanisms to determine the relevance of past and current information. Though the noise in IMU data may not wholly adhere to Markovian properties in practical applications, it usually exhibits a short correlation time scale. This means that the window of time that affects the noise at the current moment is not large. By retaining historical data while focusing more on the current moment, GRU can effectively capture the temporal dynamics of the noise, so that to accurately capture the temporal dependencies in the data.
In addition, GRU is flexible enough to handle inputs of different lengths during inference, which makes it efficient for dealing with sequences of varying lengths. The architecture of the noise net is shows in Fig. \ref{fig:noise_net}. 

The model comprises three essential parts: embedding, representation, and regression. The acceleration and gyroscope data are fed into two FC (Fully Connected) layers in the embedding layer. Then, we concatenate the resulting embedding features and input them into the representation layer. Finally, in the regression layer, we combine the representation layer's output and $V_r$ as input to obtain the final output.

In addition, since we plan to deploy the model on smartphones, we need to consider the computational resources required during inference. While maintaining high accuracy, the chosen model should be highly efficient in terms of resource consumption, including processing power and memory usage. 
The current model design ensures both accuracy and low cost. We conducted extensive evaluations to balance efficiency and accuracy, detailed experimental analysis shows in Sec. \ref{sec:exp}. 

\subsection{Input}
The input of this block contains two parts: the IMU sequential data ($Acc$ and $Gyro$) and the reference velocity ($V_r$). 

To improve computational efficiency and reduce computational resource consumption during model inference on mobile devices, we perform a 1-s feature extraction on the IMU data. For each axis of the sensor data, we extracted six time-domain features: standard deviation, maximum value, minimum value, root mean square, skewness, and kurtosis. As a result, the input dimension for both $Acc$ and $Gyro$ is 18. 

The reference velocity $V_r$ enhances the model's accuracy, which resembles the teacher forcing technique \cite{hao2022teacher} in sequence training. 
Feeding a reference $V_r$ enables the model to provide different noise compensation for different vehicle speeds. We use the velocity from the first frame within the time window as the reference velocity. During training, we utilize the velocity from GNSS while we employ the velocity estimated by the model during inference.

\section{Motion Transformation} \label{sec:mtn}
In this section, we describe our solution to solve the problem of pose transformation from phone to vehicle. We aim to eliminate the cumbersome and unstable initialization process and provide real-time output of the smartphone's relative pose to the vehicle at each moment.

\subsection{Model Design}
Inspired by Spatial Transformation Net (STN) \cite{jaderberg2015spatial}, we design a network Motion Transformation Net (MTN) to recognize the pose ($\boldsymbol{R}^v_p$) of the phone to car through acceleration measurement.

\begin{figure}[htbp]
    \centerline{\includegraphics[width=0.45\textwidth]{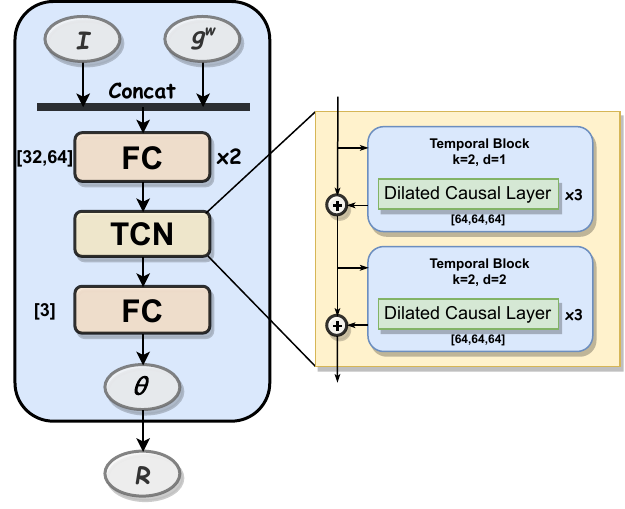}}
    \caption{The architecture of the motion transformation network. The TCN block contains 2 temporal blocks, and each tenporal block consists 3 dilated causal layers. $k$ is kernal size and $d$ is dilation factor.}
    \label{fig:mtn}
\end{figure}
To ensure the model learns the relative pose between the smartphone and the vehicle, it is crucial to focus on the changes in acceleration in both the gravity and forward directions. A temporal correlation exists in this information, i.e., the pose at the current moment is affected by previous moment.

The architecture of the MTN is shows in Fig. \ref{fig:mtn}. 
We construct the temporal model for MTN based on the Temporal Convolutional Network (TCN) \cite{bai2018empirical} architecture, as shown in Fig. \ref{fig:mtn}. TCN incorporates two core design elements for sequence learning: \textit{Dilated convolutions} and \textit{Causal convolutions}. Dilated convolutions ensure a sufficiently large receptive field, enabling efficient learning even for long sequences. Causal convolutions ensure the temporal order of sequence learning, preventing future information from leaking into the past. 

Unlike Recurrent Neural Networks (RNN), which process each time step sequentially, TCN enables parallel processing of the entire sequence. This parallel processing allows the output at the last step to comprehensively capture information from each time step in the sequence, which is advantageous for the model to learn and grasp the overall pose dynamics of the sequence.

When designing a TCN network for sequence learning, the receptive field is an essential factor to consider. By adjusting the number of temporal blocks $N_{b}$, the number of dilated convolution layers $N_{c}$ in each temporal block, and the kernel size $k$, we can flexibly control the model's receptive field. The calculation formula for the receptive field is as follows:
\begin{equation}
    Receptive\, Field = 1 + N_{c} \cdot  (k - 1) \cdot \sum_{i = 1}^{N_{b}}  2^{i-1}
\end{equation}


After the TCN blocks, we utilize FC layers to finally produce 3 euler angles.
\subsection{Input and Output}

\subsubsection{Input}
MTN takes as input the integration result of acceleration ($\boldsymbol{I}$), and the gravity acceleration reference ($\boldsymbol{g}^w$).

We divide the sequence data into one-second intervals and train the model to estimate the pose for each second.
Following the concept of pre-integration in Visual-Inertial Navigation Systems (VINS) \cite{qin2018vins}, we integrate the accelerometer data every second to reduce the sequence length and computation. For example, if the input window size is 10 seconds and the data is collected at a frequency of 50 Hz, without pre-integration, the sequence length would be $10\times 50=500$. However, pre-integration reduces the sequence length to just $10\times 1=10$. 
Furthermore, MTN's input includes a gravity acceleration reference $\boldsymbol{g}^w$, which is set to [0,0,9.81]. We concatenate the three axes acceleration integration term $\boldsymbol{I}$ with $\boldsymbol{g}^w$, resulting in a 6-dimensional input at each time step. 
\subsubsection{Output formulation}
The learning objective is to estimate the rotational Euler angles ($\boldsymbol{\theta}$), which are then transformed into a rotation matrix. We choose not to learn the rotation matrix or quaternion directly because the results provided by the model may scarcely satisfy their internal constraints, such as ensuring the rotation matrix is positively definite, or the quaternion is a unit quaternion.

Defining the three rotation angles of MTN output as $\alpha, \beta, \gamma $, the rotation matrix $\boldsymbol{R}^v_p$ can be constructed by the following formula:
\begin{equation}
    \begin{aligned}
        \boldsymbol{R}^v_p = \boldsymbol{R}_x(\alpha)\boldsymbol{R}_y(\beta)\boldsymbol{R}_z(\gamma) 
    =\begin{bmatrix}
        1 & 0            & 0             \\
        0 & \cos(\alpha) & -\sin(\alpha) \\
        0 & \sin(\alpha) & \cos(\alpha)\end{bmatrix} \\ 
        \cdot \begin{bmatrix} 
            \cos(\beta)  & 0 & \sin(\beta) \\
            0        & 1 & 0       \\
            -\sin(\beta) & 0 & \cos(\beta)
        \end{bmatrix}\cdot  \begin{bmatrix}
            \cos(\gamma) & -\sin(\gamma) & 0 \\
            \sin(\gamma) & \cos(\gamma) & 0 \\
            0 & 0 & 1
            \end{bmatrix}            
    \end{aligned}    
\end{equation}

\subsection{Data Augmentation}
To ensure effective learning of the model, we employ a data augmentation technique of randomly rotating the input data, which enhances the distribution of pose while maintaining the inherent motion characteristics. These random rotations make the model more robust and capable of handling various pose, thus improving its overall performance.

We create a random rotation matrix $\boldsymbol{R}^r$ to replicate different smartphone poses on the vehicle by randomly generating three-axis rotation Euler angles ($\alpha^r, \beta^r, \gamma^r $). This newly constructed matrix is then utilized to rotate the model's input data ($\boldsymbol{\hat{\alpha }}_p$ and $\boldsymbol{\hat{\omega }}_p$), resulting in simulated pose variations. 
\begin{equation}
    \boldsymbol{\alpha} _{in} = \boldsymbol{R}^r \boldsymbol{\hat{\alpha }}_p , \,
    \boldsymbol{\omega} _{in} = \boldsymbol{R}^r \boldsymbol{\hat{\omega }}_p
\end{equation}

To ensure consistency, we apply a uniform rotation to the accelerometer and gyroscope readings for each input window. From a model training perspective, it is important to use the same rotation matrix for each batch of data.

\section{Loss Calculation} \label{sec:loss}
In this section, we discuss the model's constraints and the loss function utilized during training. Then, we employ a sliding matching approach to mitigate the misalignment caused by potential delays in GNSS data.
\subsection{Loss Design}
For training, we use the SmoothL1Loss as loss metric, i.e.,
\begin{equation}
    L_s(x, y) = \left\{
        \begin{array}{ll}
            0.5 (x-y)^2,    &   \text{if }   {|x-y | < 1 }\\
            |x - y| - 0.5 ,   &     \text{otherwise }\\
        \end{array} 
        \right.
\end{equation}
where $x$ is the output and $y$ is the target. 
The SmoothL1Loss offers several advantages over L2. One significant advantage is its robustness to outliers, making it more stable and reliable, especially in cases where the supervised source may not provide absolute accurate ground truth. Furthermore, SmoothL1Loss exhibits a smoother gradient computation for error values around zero, leading to faster convergence than the L1 loss \cite{jadon2022comprehensive}. 

During the model training process, we not only constrain the velocity changes ($\Delta v$) but also impose constraints on the integration of velocities ($v$), which allows the model to focus on the overall motion within the window, aligning with the requirements of velocity models in practical applications. For each batch, the sequence length is $n$, and $\Delta v_i$ donate the $i$ th $\Delta v$ of this sequence, $\widehat{\Delta v}_i $ is the $i$ th target, the $\Delta v$'s loss $L_{\Delta v}$ and the $v$'s loss $L_{v}$ be like:
\begin{equation}
    L_{\Delta v} = \frac{1}{n} \sum_{i = 1}^{n}  L_s(\Delta v_i, \widehat{\Delta v}_i)
\end{equation}
\begin{equation}
    L_{v} = \frac{1}{n} \sum_{i = 1}^{n}  L_s(\int_{0}^{i} \Delta v_i \,dt , \int_{0}^{i} \widehat{\Delta v}_i \,dt)
\end{equation}

The loss $\mathcal{L}$ of this batch is:
\begin{equation}
    \mathcal{L} = \lambda  L_{\Delta v} + (1-\lambda ) L_{v}
\end{equation}
where we set the $\lambda $ to 0.7.

\subsection{Loss Matching}
When using GNSS as the primary source of supervision for motion information, temporal inconsistencies between the GNSS speed and the corresponding IMU motion data may be encountered—potential delays in the GNSS signal cause this mismatch. 
\begin{figure}[htbp]
    \centerline{\includegraphics[width=0.4\textwidth]{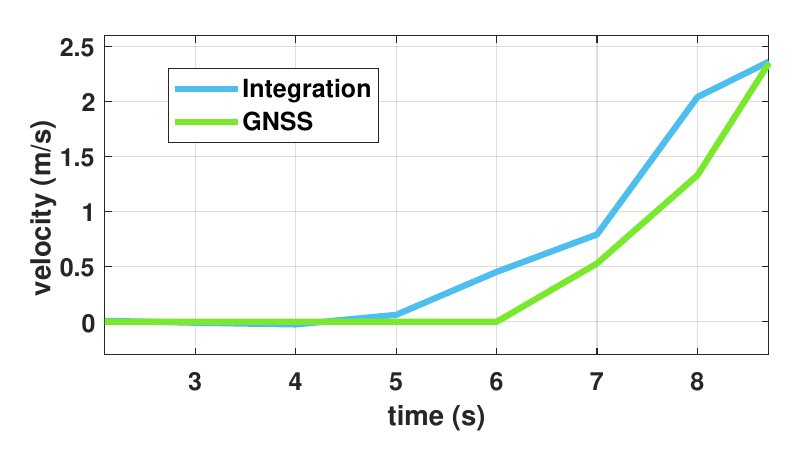}}
    \caption{An example of the delay of GNSS. The IMU (\textcolor{cyan}{blue}) records the start of a vehicle's movement before the GNSS  (\textcolor{green}{green}) provides movement information. In this particular case, the GNSS signal is delayed by approximately one second compared to the IMU signal.}
    \label{fig:gnss_delay}
\end{figure}

As shown in Fig. \ref{fig:gnss_delay}, when the vehicle starts moving, although both the IMU and GNSS sensors belong to the same mobile device and share the same timestamp, the IMU captures the vehicle's motion before the GNSS speed. If the network give out the correct estimation, the corresponding speed in a delay will likely not match the target, inducing a high error penalty. So that we adjust to match the target. However, the duration of this delay is not fixed, making it challenging to align the data directly using a specific value during the data processing stage. Inspired by Godard's work \cite{godard2019digging}, we dynamically align this uncertain delay during the loss computation stage.

\begin{figure}[htbp]
    \centerline{\includegraphics[width=0.5\textwidth]{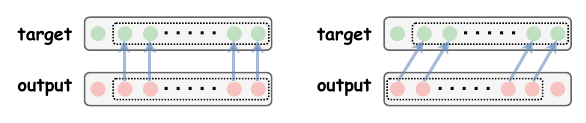}}
    \caption{Two alignments for loss matching. }
    \label{fig:loss_match}
\end{figure}

We utilize a sliding window to achieve a more accurate match between the output and the target.
As mentioned before, the output sequence is one-second interval. If there is no misalignment between the output and the target, we can calculate the loss directly. However, if there is a delay in the target, we compute the loss using the output from the previous time step and the target at the current time step. We can gradually calculate the loss with shifting misalignment by sliding the window backward.
However, in reality, the delay in GNSS signals is limited. Although we have not accurately measured the distribution of time delays, based on observations, this delay usually is at most one second when the GNSS signal is strong. Therefore, when calculating the loss, we only consider two sets: one with aligned timestamps and another with a one-second offset.
As shown in Fig. \ref{fig:loss_match}, we calculate the losses of the two sets of data respectively, then use the minimum of them:
\begin{equation}
    \mathcal{L} = min(\mathcal{L}(x_{[2:n]}, y_{[2:n]}), \mathcal{L}(x_{[1:n-1]}, y_{[2:n]}))
\end{equation}
\section{Experiments} \label{sec:exp}
In this section, we validate our framework's effectiveness through experiments. We introduce the dataset, implementation details, and evaluation metrics. Comparisons with baseline methods are made for performance and generalizability. We analyze various temporal models and conduct ablation experiments to analyze component contributions.
\subsection{Dataset}
Our dataset is sourced from our crowdsourcing data platform, where we incentivize drivers to download a data collection app and upload their driving data. All data undergoes anonymization and does not contain any user information. The dataset includes driving data from over 300 drivers, with a total duration of approximately 200 hours. The data collection period spans from March 2023 to December 2023. We only extract the IMU and GNSS data from the dataset, excluding device information. Additionally, the phone's pose inside the vehicle during data collection is unknown.

\subsection{Implementation}

\subsubsection{Data split} \label{subsec:datasplit}
When preparing the training and testing data for the model, we split the dataset in two ways: traditional split and trajectory split.

Traditional split loads all the data and divids them into batches, then splits them into training, validation, and testing sets in a ratio of 7:1:2.

Trajectory split, on the other hand, prioritizes dividing the data based on trajectories. We randomly selected 20\% of the trajectories from the entire dataset as the testing data. Then, we shuffled the remaining data and divided it into training and validation data in an 8:2 ratio. This partitioning approach ensures the presence of devices and poses in the testing data that may not involved in the model training, thereby validating the model's generalization performance.

In subsequent experiments, we use the trajectory split if we do not explicitly state the use of the traditional split dataset.
\subsubsection{Training details}
In the implementation, we use PyTorch to train our model on the machine learning platform with one GeForce GTX 2080 Ti GPU, an Intel i7 CPU, and 40G RAM. We set the batch size to 512 and used Adam Optimizer. We adopt a schedular of learning rate initialized in 1e-3 and decay it following cosine annealing strategy \cite{loshchilov2017decoupled}, and we set an early stopping. In addition, we implement Exponential Moving Average (EMA) \cite{ema} and Automatic Mixed Precision (AMP) \cite{micikevicius2017mixed} to improve model's robustness and training speed.
\subsubsection{Implementation on Smartphone}
To deploy the model on a smartphone, we leverage the ONNX-Runtime platform. After training the model on the server, we convert the model into an ONNX format, which can directly load and execute inference on Android smartphones with ONNX-Runtime. 

\subsection{Metrics}
We evaluate the model's accuracy by measuring the errors in velocity and distance. The learning target of the model in velocity change within 1s. Vehicle speed is the first order integration, and by assessing the error between the inferred velocity and the actual velocity for each second, we can intuitively judge the model's precision. However, as mentioned before, the velocity provided by GNSS may not perfectly match the actual vehicle motion. Therefore, we additionally evaluate the distance obtained through velocity integration. Integration can amplify error accumulation and more clearly reflect the model's accuracy. We chose integration durations of 30 and 60 seconds, sufficient to cover most scenarios where GNSS signals fail.
Our evaluation metrics include the mean and 80th percentile of the error. These metrics provide a comprehensive assessment of the model's performance.

\subsection{Comparison with Others}
\subsubsection{Baselines}
We compare our method against the following baseline methods to demonstrate its superiority in vehicle speed estimation via smartphone. 

\begin{itemize}
    \item AI-IMU \cite{brossard2020ai}: AI-IMU is built upon the foundation of the EKF, which incorporates two key elements: the Kalman filter and the dynamic adjustment of noise covariance matrices through deep neural networks. This approach includes a state-space model and simple car movement assumptions. By utilizing the Kalman filter, this method leverages the kinematic assumptions and melds them with statistical analysis of the learning model's results.
    \item VeTorch \cite{gao2021glow}: Vetorch utilizes TCN to encode IMU data for velocity estimation. In order to ensure that the smartphone and vehicle coordinate systems are properly aligned, the acceleration on the horizontal plane is first calculated using gravity acceleration. Next, the inertial sequence is extracted while the vehicle is in motion, and PCA is applied to decompose the forward direction vector. Finally, Vetorch constructs orthogonal three-dimensional unit vectors by combining the gravity and forward directions, yielding a reliable rotation matrix. To combat generalization challenges, Vetorch suggests a personalized approach using federated learning, which allows for the training of individualized models for each user.
    \item DeepTrack \cite{tong2022vehicle}: DeepTrack follows a similar process to Vetorch, utilizing PCA for pose estimation, TCN for temporal data encoding, and personalized training to optimize user models. The main differences lie in their model structures, and DeepTrack employs features rather than raw data during the training process.
    \item DeepVIP \cite{zhou2022deepvip}: DeepVIP leverages geomagnetic information to achieve alignment between the smartphone and vehicle coordinate systems. Additionally, it incorporates magnetic signals during motion sequence learning. Notably, DeepVIP introduces two network architectures: an LSTM-based network (DeepVIP-L) achieving higher accuracy and a MobileNet-based network (DeepVIP-M) becoming more compact. In our replication, we utilize the DeepVIP-L, which offers higher accuracy.
\end{itemize}

The variations in implementation, availability of code, and data in different methodologies make it challenging to ensure a fair and equal comparison among them. However, it is essential to strive towards maintaining the core design principles of the original works when implementing. We try to preserve the essence of each approach and compare them on an equal footing.

\begin{table*}[htbp]
    \centering
    \caption{The accuracy comparison of different method}
    \begin{tabular}{lrrrr|rrrr}
        \toprule
        \multirow{3}[6]{*}{} & \multicolumn{4}{c|}{\textbf{30s}} & \multicolumn{4}{c}{\textbf{60s}} \\
        \cmidrule{2-9}      & \multicolumn{2}{c}{\textbf{Velocity}} & \multicolumn{2}{c|}{\textbf{Distance}} & \multicolumn{2}{c}{\textbf{Velocity}} & \multicolumn{2}{c}{\textbf{Distance}} \\
        \cmidrule{2-9}      & \multicolumn{1}{c}{\textbf{MAE}} & \multicolumn{1}{c}{\textbf{80\%}} & \multicolumn{1}{c}{\textbf{MAE}} & \multicolumn{1}{c|}{\textbf{80\%}} & \multicolumn{1}{c}{\textbf{MAE}} & \multicolumn{1}{c}{\textbf{80\%}} & \multicolumn{1}{c}{\textbf{MAE}} & \multicolumn{1}{c}{\textbf{80\%}} \\
        \midrule
        \textbf{AI-IMU} \cite{brossard2020ai} & 4.33  & 6.74  & 48.02 & 78.77 & 6.42  & 10.22 & 144.06 & 245.47 \\
        \textbf{Vetorch} \cite{gao2021glow} & 5.05  & 6.94  & 68.41 & 93.16 & 5.99  & 8.27  & 162.31 & 227.16 \\
        \textbf{DeepVIP} \cite{zhou2022deepvip} & 2.91  & 3.8   & 33.46 & 45.32 & 4.47  & 5.56  & 102.93 & 138.61 \\
        \textbf{DeepTrack} \cite{tong2022vehicle} & 1.98  & 2.93  & 23.47 & 35.03 & 2.51  & 3.63  & 58.56 & 91.68 \\
        \textbf{DVSE} & \textbf{1.78} & \textbf{2.76} & \textbf{20.04} & \textbf{31.83} & \textbf{2.35} & \textbf{3.62} & \textbf{50.84} & \textbf{83.65} \\
        
        \bottomrule
        \end{tabular}%
    \label{tab:accuracy}%
\end{table*}%

\subsubsection{Accuracy}
We conducted a comparison of various baseline methods and our approach for accuracy. For AI-IMU, after adjusting many initialization parameters, it takes some time to converge, so we remove the segments that converge during evaluation. Beside AI-IMU, all methods underwent the same training process and configuration, with a 10-second input sequence length. VeTorch and DeepTrack require data projection during application and initialization using the first 100 seconds of the test trajectory to ensure complete performance. We utilized the GNSS signal to identify stationary and forward data segments and calculate the rotation matrix between the phone and vehicle systems. DeepVIP, on the other hand, performs coordinate system transformation every second using magnetometer and gravity acceleration data. Our approach does not require any initialization steps. 

Table \ref{tab:accuracy} presents the accuracy results of each baseline method and DVSE on our dataset. Our approach achieves the highest accuracy without the need for initialization parameters. DeepTrack also shows good results with a favorable initialization. However, the unstable magnetometer affect the accuracy of DeepVIP, resulting in significant degradation in some cases. VeTorch exhibit a weaker fitting ability for temporal data due to its immature network structure. The AI-IMU's dependence on initialization parameters is noticeable, and the time taken for convergence is unsteadiness. Even after removing the stage of convergence, the performance remains inadequate. Furthermore, VeTorch and DeepVIP use raw data as input, resulting in high-dimensional feature inputs and slower inference speed. 

\subsubsection{Generalization}
\begin{figure}[htbp]
    \centerline{\includegraphics[width=0.45\textwidth]{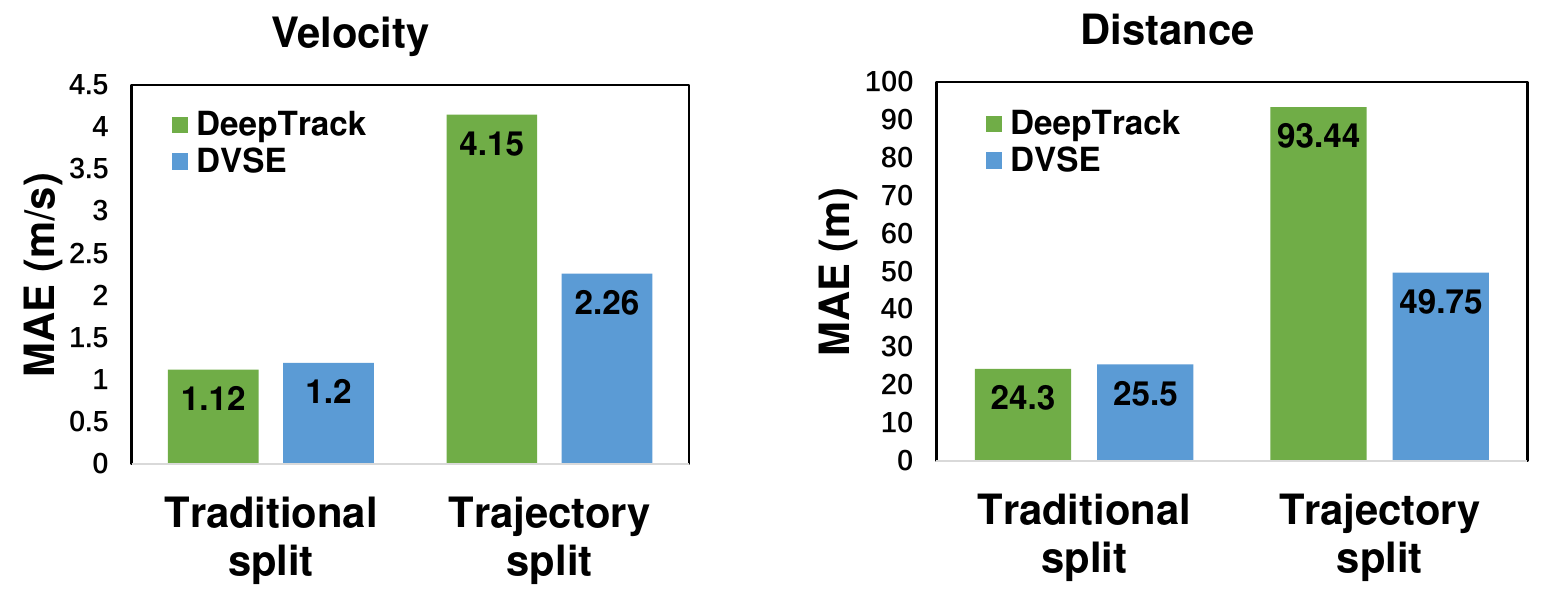}}
    \caption{Comparison of DeepTrack and DVSE to different dataset split method. }
    \label{fig:gen}
\end{figure}

We conducted experiments to validate the generalization performance of models by comparing its performance using different data split methods. We trained and tested models using Traditional split and Trajectory split (described in Sec. \ref{subsec:datasplit}) separately. We solely compared our mehthod to DeepTrack, which also considered the generalization. The experiment disregard any initialization steps or subsequent optimization operations. The experimental results are illustrated in Fig. \ref{fig:gen}.

Based on the traditional data split method, DeepTrack performed well, achieving a cumulative average velocity error of 1.12 m/s over 60 seconds. Our method, on the other hand, got a velocity error of 1.2 m/s, slightly inferior to DeepTrack. Both methods demonstrated strong fitting capabilities for velocity estimation using IMU sequences. However, when tested on the trajectory split dataset, DeepTrack's performance deteriorated significantly, with a velocity error of 4.15 m/s, representing a degradation of 370\% compared to the traditional split dataset. In contrast, our method had an error of 2.26 m/s, representing a degradation of 180\%. These findings indicate that DeepTrack struggles to adapt well when confronted with trajectories in the test data that were absent in the training data. In contrast, our method exhibits better adaptability to unknown data.

Indeed, we desire plug-and-play functionality when using deep learning-based methods rather than requiring cumbersome initialization processes. We cannot expect users to perform extensive operations in practical applications, such as "drawing figure eight" before each use. While the personalized approaches employed by Vetorch and DeepTrack significantly enhance the model's generalization capabilities, the current deployment cost of customized training is high, especially for application providers who cannot afford to allocate substantial time and computational resources for training personalized models. 

When it comes to estimating vehicle dynamics through smartphones, two crucial factors impacting generalization performance are noise and smartphone placement disparities. Our approach focuses on augmenting data of smartphone poses rather than relying on a vast amount of high-quality data, which also effectively enhances the model's generalization capabilities. However, managing discrepancies in noise remains a significant obstacle that necessitates continued investigation. Moving forward, we will continue to explore more efficient generalization models to address this challenge.

\subsection{Comparison of Network}
We compared the performance of mainstream temporal learning networks in our noise compensation block. 
Considering that our network will ultimately run on smartphones, we compared the accuracy and evaluated the memory usage and average inference time during runtime. 
The experiment solely involved replacing the noise compensation network (GRU in Fig. \ref{fig:noise_net}) with different networks while keeping the environment consistent. 
We gradually reduce different networks' parameter count and balance their performance while ensuring accuracy. 
Among them, LSTM and GRU have the same configuration, and both are one layer and have 128 hidden units. On maintaining a receptive field of 10, TCN sets the number of convolution kernels for dilated causal convolution to 128; Attention embeds the input data and encodes it using an encoder with four heads and hidden layers' dimension set to 64.
The result shows in Table \ref{tab:network}.

\begin{table}[htbp]
    \centering
    \caption{Comparison of different network}
      \begin{tabular}{|c|c|c|c|c|c|}
      \hline
      \multirow{2}{*}{\textbf{}} 
      & \multicolumn{2}{c|}{\textbf{Velocity}} 
      & \multicolumn{2}{c|}{\textbf{Distance}} 
      & \multicolumn{1}{c|}{\multirow{2}{*}{\makecell[c]{\textbf{RAM}}}} \\
  \cline{2-5}          & \textbf{MAE} & \textbf{80\%} & \textbf{MAE} & \textbf{80\%} &         \\
      \hline
      \textbf{LSTM} & 2.74  & 4.06  & 59.45 & 90.43   & 733K    \\
      \hline
      \textbf{GRU} & 2.35  & 3.6   & 50.84 & 83.65  & 572K    \\
      \hline
      \textbf{TCN} & 2.48  & 3.61  & 54.48 & 86.92  & 719K    \\
      \hline
      \textbf{Attention} & 2.74  & 4.1   & 62.78 & 99.21  & 3473K    \\
      \hline
      \end{tabular}%
    \label{tab:network}%
\end{table}%
  
\begin{figure}[htbp]
    \centerline{\includegraphics[width=0.45\textwidth]{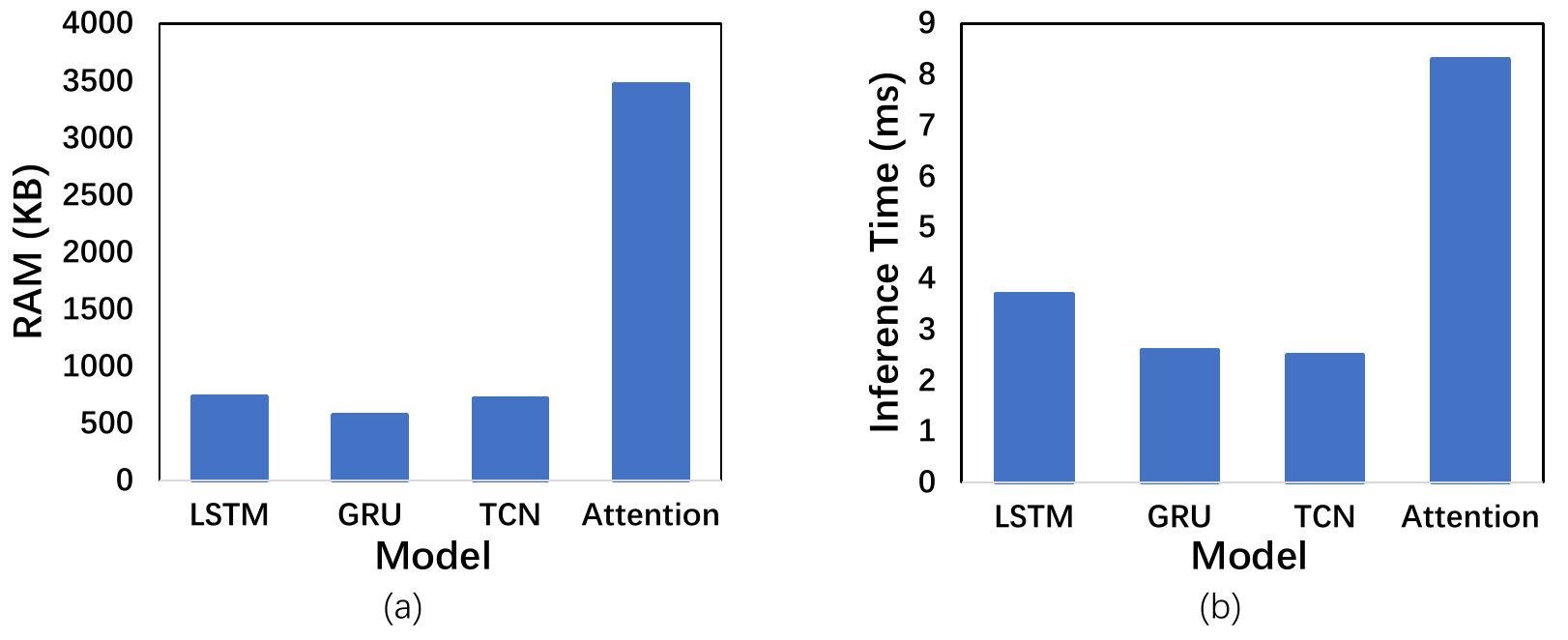}}
    \caption{The RAM and inference time of different networks when deployed in smartphone.}
    \label{fig:ram}
\end{figure}

Among the different networks evaluated, the GRU network demonstrated superior performance. When comparing velocity and distance, the GRU network yielded the lowest MAE and 80th percentile error. The TCN network closely followed while the LSTM and Attention networks exhibited slightly lower performance. However, they still demonstrated satisfactory accuracy, indicating that all networks are capable of effectively learning temporal tasks. The Attention network's suboptimal performance can be attributed to the short sequence length, which constrained its ability to fully capitalize on its strengths in learning long sequences. In the deployment comparison experiment, the GRU network continued to outperform the others. It achieved the minimum memory usage with the fewest parameters and had the fastest inference speed.
The deployment results are visualized as histograms, as shown in Figure \ref{fig:ram}. 
The test phones are Samsung S9, Huawei P40, Oppo Reno 10 Pro, and we take the average results.

Although the GRU network boasts the highest accuracy, its superiority over other networks is not substantial. By tweaking the model architecture and parameters, comparable levels of accuracy can be attained across various models. Nevertheless, taking deployment into account, we prefer to opt for a model that consumes less memory and offers quicker inference speed. Hence, we have decided to utilize the GRU network as the foundation of our noise compensation network.

\subsection{Ablation}
To demonstrate the effectiveness of different modules, we conducte a series of ablation experiments on our dataset. Four varients are considered:
\begin{itemize}
    \item Remove Data augmentation (w/o Data Aug).
    \item Remove Noise compensation network (w/o Noise).
    \item Remove Motion transformation network (w/o MTN).
    \item Remove Loss matching (w/o Loss Match)
\end{itemize} 
\begin{table}[htbp]
    \centering
    \caption{Ablation results of our methods (60 seconds)}
      \begin{tabular}{|c|c|c|c|c|}
      \hline
      \multirow{2}[4]{*}{} & \multicolumn{2}{c|}{\textbf{Velocity}} & \multicolumn{2}{c|}{\textbf{Distance}} \\
  \cline{2-5}          & \textbf{MAE} & \textbf{80\%} & \textbf{MAE} & \textbf{80\%} \\
      \hline
      \textbf{Base} & 2.35  & 3.62  & 50.84 & 83.65 \\
      \hline
      \textbf{w/o Noise} & 3.43  & 5.57  & 75.9  & 129.16 \\
      \textbf{w/o MTN} & 3.65  & 5.27  & 79.71 & 114.82 \\
      \textbf{w/o Data Aug} & 3.72  & 6.21  & 84.06 & 148.41 \\
      \textbf{w/o Loss Match} & 2.79  & 4.07  & 63.8  & 97.85 \\
      \hline
      \end{tabular}%
    \label{tab:ablation}%
\end{table}%

Ablation experiments are conducted in the same environment and training settings, and the results are presented in Table \ref{tab:ablation}. The table illustrates the effectiveness of each module. 
Compare to the base, for w/o noise, velocity MAE and 80 quantile decrease by 1.08 m/s ($\downarrow\!\!29\%$) and 1.95 m/s ($\downarrow \!\!35\%$), distance MAE decrease by 25 m ($\downarrow\!\! 33\%$). The noise compensation network effectively captures the differences between sensor measurements and actual motion, ensuring that the model goes beyond the simple integration of measurements after pose rotation. 
For w/o MTN, velocity MAE reduces by 1.3 m/s ($\downarrow \!\!35.6\%$) and 28 m ($\downarrow \!\!35.4\%$) on distance. By aligning the smartphone coordinate system with the vehicle coordinate system, the MTN avoids errors in estimating vehicle speed due to smartphone attitude issues. 
For w/o Data Aug, velocity MAE reduces by 1.4 m/s ($\downarrow \!\!37.6\%$) and 34 m ($\downarrow \!\!40.5\%$) on distance. Data augmentation allows for simulating arbitrary smartphone attitudes on the vehicle, enhancing the model's generalization performance. Consequently, it demonstrates a noticeable improvement in accuracy on the trajectory-split dataset. 
w/o Loss Match shows the reduction in velocity and distance MAE by roughly 0.34 m/s ($\downarrow \!\!12.2\%$) and 13 m ($\downarrow \!\!20.4\%$). The loss matching module helps alleviate the problem of vehicle dynamics mismatch between IMU observations and GNSS information caused by latency, improving the model's learning accuracy and reducing the threshold for data selection, significantly increasing the training data.
  
Further, each module contributes to improving accuracy to varying extents. Among them, the data augmentation module shows the most significant improvement in accuracy, attributed to the limited distribution of training data. The experimental group uses trajectory-split datasets, and the test set contains unknown smartphone poses in the training set, which indicates that the model lacks the ability to handle unknown attitudes accurately, significantly affecting the speed model's accuracy in practical applications. Although this problem can be addressed by collecting sufficient data, it raises significant challenges in data acquisition. Data augmentation expands the pose distribution, ensuring that the model fits the motion observations resulting from arbitrary smartphone rotations during training, thereby enhancing the model's generalization performance. The noise compensation and motion transformation modules, as the core of the model, also contribute significantly to improving accuracy. While the loss matching module demonstrates some improvement in accuracy, it may not be as pronounced. This suggests that our approach is effective and also indicates that GNSS latency is not always present.
However, performing sliding matching at the second level may not always precisely align the timestamps between IMU and GNSS. In future work, we will explore better matching approaches to address this issue.

\section{Conclusion} \label{sec:conclusion}
In this paper, we present a novel deep learning-based method to estimate vehicle speed with only smartphone's inertial data. We split the learning process into a motion transformation module and a noise compensation module to simplify the learning logic of the model and use data augmentation technology to expand the pose distribution of the training data. In addition, we optimize the matching loss for the training task using GNSS as the supervision source. Finally, we design a prototype system and validate it on our real-world crowdsourcing data. The results show that our method outperforms other methods regarding accuracy and generalization performance.

\bibliographystyle{IEEEtran}
\bibliography{mybib}


\newpage






\vfill

\end{document}